\documentclass[letterpaper, 10 pt, conference]{ieeeconf}  % Comment this line out if you need a4paper

\IEEEoverridecommandlockouts                              % This command is only needed if 
\usepackage{caption}
\usepackage{stfloats}
\usepackage{graphics} % for pdf, bitmapped graphics files
\usepackage{epsfig} % for postscript graphics files
\usepackage{mathptmx} % assumes new font selection scheme installed
\usepackage{times} % assumes new font selection scheme installed
\usepackage{amsmath} % assumes amsmath package installed
\usepackage{amssymb}  % assumes amsmath package installed
\usepackage{multicol}
\usepackage{multirow}
\usepackage{booktabs} 
\usepackage{bbding}
\usepackage{tabularx}
\usepackage{cite}
\usepackage{hyperref}

\title{\LARGE \bf
Complementing Onboard Sensors with Satellite Maps: A New Perspective for HD Map Construction
}

\author{Wenjie Gao$^{1}$, Jiawei Fu$^{2}$, Yanqing Shen$^{1}$, Haodong Jing$^{1}$, Shitao Chen$^{1\dag}$, Nanning Zheng$^{1}$% <-this % stops a space
\thanks{This work is supported by the National Key R\&D Program of China (Grant No. 2022YFB2502900) and the National Natural Science Foundation of China (Grant No. 62088102).}
\thanks{$^{1}$W. Gao, Y. Shen, H. Jing, S. Chen and N. Zheng are with National Key Laboratory of Human-Machine Hybrid Augmented Intelligence, National Engineering Research Center for Visual Information and Applications, and Institute of Artificial Intelligence and Robotics, Xi'an Jiaotong University, Shaanxi 710049, P.R. China. \{{\tt\small gaowenjie999}, {\tt\small qing1159364090}, {\tt\small jinghd}\} {\tt\small @stu.xjtu.edu.cn}; \{{\tt\small chenshitao}, {\tt\small nnzheng}\}{\tt\small @mail.xjtu.edu.cn}}%
\thanks{$^{2}$J. Fu is with The Chinese University of Hong Kong, Shatin, Hong Kong. {\tt\small jwfu@cse.cuhk.edu.hk}}%
\thanks{S. Chen$\dag$ is with the corresponding author.}%
}

\begin{document}

\maketitle
\thispagestyle{empty}
\pagestyle{empty}

%%%%%%%%%%%%%%%%%%%%%%%%%%%%%%%%%%%%%%%%%%%%%%%%%%%%%%%%%%%%%%%%%%%%%%%%%%%%%%%%
\begin{abstract}

High-definition (HD) maps play a crucial role in autonomous driving systems. Recent methods have attempted to construct HD maps in real-time using vehicle onboard sensors. Due to the inherent limitations of onboard sensors, which include sensitivity to detection range and susceptibility to occlusion by nearby vehicles, the performance of these methods significantly declines in complex scenarios and long-range detection tasks. In this paper, we explore a new perspective that boosts HD map construction through the use of satellite maps to complement onboard sensors. We initially generate the satellite map tiles for each sample in nuScenes and release a complementary dataset for further research. To enable better integration of satellite maps with existing methods, we propose a hierarchical fusion module, which includes feature-level fusion and BEV-level fusion. The feature-level fusion, composed of a mask generator and a masked cross-attention mechanism, is used to refine the features from onboard sensors. The BEV-level fusion mitigates the coordinate differences between features obtained from onboard sensors and satellite maps through an alignment module. The experimental results on the augmented nuScenes showcase the seamless integration of our module into three existing HD map construction methods. The satellite maps and our proposed module notably enhance their performance in both HD map semantic segmentation and instance detection tasks. Our code will be available at \href{https://github.com/xjtu-cs-gao/SatforHDMap}{\textit{https://github.com/xjtu-cs-gao/SatforHDMap}}.

\end{abstract}

%%%%%%%%%%%%%%%%%%%%%%%%%%%%%%%%%%%%%%%%%%%%%%%%%%%%%%%%%%%%%%%%%%%%%%%%%%%%%%%%
\section{Introduction}

As an essential component in autonomous driving systems, high-definition (HD) maps contain precise geographic information and rich semantic details of map elements such as pedestrian crossings, lane dividers, and road boundaries. This information within HD maps enables ego-vehicle to locate itself within the road network, as well as provides route and navigation information for downstream prediction and motion planning modules. Conventional manual-based offline annotation approaches confront widespread implementation challenges, primarily due to their high labor costs. Recent research\cite{liHdmapnetOnlineHd2022, liuVectormapnetEndtoendVectorized2022, liaoMapTRStructuredModeling2022} attempted to construct HD maps online using vehicle onboard sensor data and achieved good performance. However, we observe that these methods are affected by vehicle's surrounding environment and detection range due to the inherent limitations of onboard sensors, including weak detection for long-range objects and vulnerability to occlusion by neighboring vehicles. 

Our primary insight underscores the potential augmentation of HD map construction by complementing onboard sensors with cloud-based satellite maps. Rich cloud-based information can be easily accessed for vehicles during the driving process, including roadside data, previous bird’s-eye view (BEV) information\cite{xiong2023neural}, and satellite maps of driving area. Among this information, satellite maps provide distinctive advantages. Firstly, satellite maps have the ability to cover the driving area, thus providing long-range information. Secondly, the top-down perspective provided by satellite maps is less prone to obstruction by other vehicles and can complement the perspective view of onboard sensors. Finally, given that most existing methods convert the perspective view into BEV before calculation, the satellite map information can be seamlessly integrated with these methods due to its top-down perspective. 

\begin{figure*}[!ht]
    \setlength{\abovecaptionskip}{0cm}

    \centering
    \includegraphics[width=1.8\columnwidth]{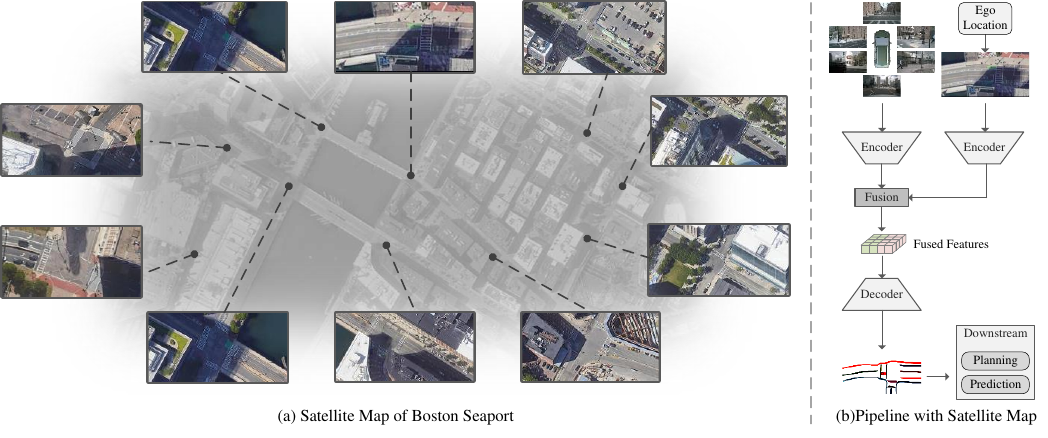}
    \caption{(a) Satellite maps provide comprehensive insights into the surrounding region. (b) The satellite map tile of ego location can be integrated into the current HD map construction pipeline to complement onboard sensors.}
    \label{satellite}
    % \vspace{-1em}
\end{figure*}

In this study, we explore the seamless and efficient integration of satellite maps into existing HD map construction methods, as illustrated in Fig. \ref{satellite}. Our initial work involves generating satellite map tiles corresponding to each sample in nuScenes\cite{caesar2020nuscenes}, complementing the nuScenes dataset. Considering that nuScenes utilizes a custom coordinate system, we establish a coarse coordinate transformation equation and generate satellite map tile on each sample's pose in nuScenes. The next challenge is \textbf{how to make full use of the satellite maps}. There are coordinate deviations after coarse coordinate transformation and obstruction of satellite maps by trees in minor scenarios, hindering the integration of satellite maps. To address the issues, we propose a hierarchical fusion module comprising feature-level fusion and BEV-level fusion. The former utilizes a masked cross-attention mechanism to refine features from onboard sensors using satellite maps, in which the mask is designed to avoid interference from irrelevant information in satellite maps and reduce unnecessary interactions. The latter uses an alignment module to alleviate the impact of coordinate deviations.

% experiment
We perform comprehensive experiments to assess the performance of various fusion methods and validate the efficacy of satellite maps. The results show that even the simplest fusion method, such as concatenation directly, can significantly enhance the performance of the baseline model, indicating that satellite maps are an effective complement to onboard sensors. We further conduct experiments to integrate our proposed hierarchical fusion module into three HD map construction methods. The results showcase remarkable performance improvements of +20.8 mIoU for HDMapNet\cite{liHdmapnetOnlineHd2022}, +7.9 mAP for VectorMapNet\cite{liuVectormapnetEndtoendVectorized2022}, and +2.3 mAP for MapTR\cite{liaoMapTRStructuredModeling2022} in both map semantic segmentation and instance detection tasks.

To summarize, our contributions include the following:

\begin{itemize}
    \item We proactively explore the significance of satellite maps in HD map construction and release a complementary satellite map dataset for nuScenes, providing a new perspective for future HD map construction research.
    \item We propose a hierarchical fusion module to facilitate better fusion between satellite map and onboard sensor information. The feature-level fusion utilizes relevant information from satellite maps to enhance features from onboard sensors, while the BEV-level fusion mitigates the impact of coordinate offsets before concatenation.
    \item We integrate our module into three existing HD map construction methods and demonstrate significant improvement, particularly in long-range map construction.
\end{itemize}

\section{Related Works}

\captionsetup[figure]{name={Fig.},labelsep=period,singlelinecheck=off} 

\subsection{HD Map Construction}

HD maps are an indispensable component in autonomous driving systems. Conventional HD maps are constructed by SLAM-based methods\cite{zhang2014loam, shan2020lio, shan2018lego, mur2017orb}. Recently, some methods \cite{Peng_2023_WACV, zhouCrossviewTransformersRealtime2022, philion2020lift, dongSuperFusionMultilevelLiDARCamera2022, wang2023lidar2map} perform segmentation in a rasterized BEV space to obtain drivable areas or map elements. This representation, however, lacks instance-specific information and is proved incompatible with downstream modules\cite{gao2020vectornet}. Instead, recent works \cite{qiao2023end, shinInstaGraMInstancelevelGraph2023, liaoMapTRStructuredModeling2022, liuVectormapnetEndtoendVectorized2022, liHdmapnetOnlineHd2022, xiong2023neural, xie2023mv, zhang2023online} represent map elements as instances composed of individual points. To construct vectorized HD maps, HDMapNet\cite{liHdmapnetOnlineHd2022} groups pixel-wise segmentation results with heuristic post-processing, which requires a significant amount of computations. VectorMapNet\cite{liuVectormapnetEndtoendVectorized2022}, InstaGraM\cite{shinInstaGraMInstancelevelGraph2023} and MapTR\cite{liaoMapTRStructuredModeling2022} achieve end-to-end map element detection. The aforementioned methods rely exclusively on data from onboard sensors as inputs. A similar study NMP\cite{xiong2023neural} proposes a new paradigm by acquiring historical BEV information of the ego-vehicle or other vehicles from cloud or local storage as prior knowledge. In contrast, satellite maps provide heightened accessibility and our proposed fusion module demonstrates superior performance in fusion. 

\subsection{Multi-Sensor Fusion}

Multi-sensor fusion has been a prominent research area in the field of autonomous driving. Presently, prevalent sensor fusion methods can be categorized into Transformer-based methods\cite{dongSuperFusionMultilevelLiDARCamera2022, bai2022transfusion, pang2023transcar, chittaTransfuserImitationTransformerbased2022, 10.1007/978-3-031-20074-8_36, zhangCatdetContrastivelyAugmented2022, wang2021pointaugmenting} and concatenation-based methods\cite{liu2023bevfusion, liang2022bevfusion, dongSuperFusionMultilevelLiDARCamera2022, meyer2019sensor}. The fundamental paradigm of Transformer-based methods involves mapping two types of features into a shared feature space, constructing queries, keys, and values, and subsequently utilizing attention mechanisms to facilitate fusion. Transfuser\cite{chittaTransfuserImitationTransformerbased2022} concatenates LiDAR features and camera features, leveraging standard self-attention modules for fusion. AutoAlignV2\cite{10.1007/978-3-031-20074-8_36} introduces a multi-layer deformable cross-attention network to aggregate features from distinct modalities. The primary challenge encountered by concatenation-based methods pertains to the alignment of data originating from disparate sensors. Recent BEVFusion works\cite{liu2023bevfusion, liang2022bevfusion} use fully convolution layers with few residual blocks to compensate for such localized misalignments. Nevertheless, applying the above-mentioned methods to integrate satellite map into existing approaches directly yields suboptimal results. Therefore, we combine the unique attributes of satellite map and drew inspiration from these methods to design a hierarchical fusion module.

\begin{figure*}[t]

    \centering
    \includegraphics[width=2\columnwidth]{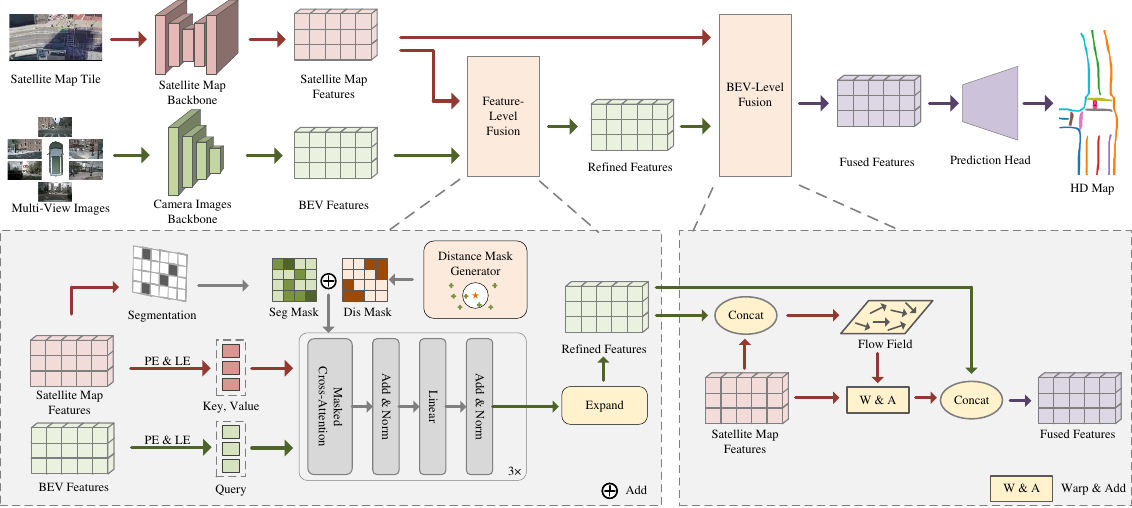}
    \caption{Framework overview. PE stands for patch embedding and position embedding, LE stands for linear embedding. The green arrows represent the information flow from onboard sensors. The red arrows represent the information flow from satellite maps. Our framework utilizes two branches to extract features from multi-view images and satellite map tiles, respectively. A hierarchical fusion module, comprising feature-level fusion and BEV-level fusion, is designed to fuse the two features. The final task head is used to generate the HD maps from the fused features.}
    \label{system_overview_figure}
    \vspace{-1em}
\end{figure*}

\section{Complementary Dataset for nuScenes} \label{dataset section}

The nuScenes\cite{caesar2020nuscenes} dataset is widely used in the field of autonomous driving, and most existing methods of HD map construction have been validated on it. It encompasses the entire suite of sensors for autonomous vehicles, including 6 cameras, 5 radars, 1 LiDAR, and GPS \& IMU, and is composed of 1000 scenes lasting 20s duration, collected across four districts. However, nuScenes only incorporates information from onboard sensors. Therefore, we obtain satellite maps of the four districts as a complementary dataset. The details are shown in Table \ref{table-dataset}.

\begin{table}[h]
\centering
\begin{scriptsize}
\caption{Details of complement dataset for four districts. Resolution stands for the resolution of satellite map tiles.}
\label{table-dataset}
\begin{tabular}{c|ccc}
\toprule
 District & Area Size & Samples & Resolution \\
\midrule
Boston Seaport & $(2200m, 3500m)$ & 22266& (137, 273) \\
Singapore’s One North & $(1700m, 2000m)$ & 8143 & (102, 202) \\
Queenstown & $(3100m, 3600m)$ & 6604 & (102, 202) \\
Holland Village & $(2600m, 2600m)$ & 3548 & (102, 202) \\
\bottomrule
\end{tabular}
\end{scriptsize}
\vspace{-2em}
\end{table}

Here we describe the process of generating satellite map tiles for each sample. Initially, the highest resolution satellite maps of four districts in nuScenes are downloaded from Google Maps. Next, we utilize a keypoint alignment method that selects the coordinates of five landmarks for alignment, establishing the coarse transformation equation between the nuScenes coordinate system and the satellite map coordinate system. After coarse alignment, the coordinate deviation is within $2m$. Following this, the corresponding satellite map region is acquired based on the position and orientation of each sample. The satellite maps are then sliced into the size of $(30m, 60m)$, which matches the configuration of most HD map construction methods. Ultimately, we retrieve the tiles of these regions using the bilinear interpolation method.

We release the satellite map tiles as a complementary dataset for nuScenes, which is available at \href{https://www.kaggle.com/datasets/wjgao0101/satfornuscenes}{\textit{https://www.kaggle.com/datasets/wjgao0101/satfornuscenes}}.

\section{Satellite Map Fusion Framework}

In Section \ref{dataset section}, we present a complementary dataset of satellite maps for nuScenes. In this section, we introduce a framework based on a hierarchical fusion module for integrating the satellite maps into existing HD map construction methods, as shown in Fig. \ref{system_overview_figure}.

\subsection{Overall Architecture}

% image branch 需要修改
The general paradigm for HD map construction methods follows an encoder-decoder architecture, in which the encoder transforms data from onboard sensors, such as camera images, into BEV features and the decoder constructs HD maps from the extracted BEV features. 

Within our framework, we utilize two distinct branches to extract features from camera images and satellite maps, respectively. The camera images branch follows HDMapNet, VectorMapNet, and MapTR \cite{liHdmapnetOnlineHd2022, liuVectormapnetEndtoendVectorized2022, liaoMapTRStructuredModeling2022} to take the camera images input $\mathcal{I}_{img} \in \mathbb{R}^{6 \times H_i \times W_i \times 3}$ from onboard sensors. Various methods, including MLP, Inverse Perspective Mapping (IPM)\cite{mallot1991inverse} can be utilized to transform the images from perspective view to BEV view to obtain the BEV features $\mathcal{F}_{bev} \in \mathbb{R}^{H \times W \times C}$, where $H=100$, $W=200$, and $C=64$. For the satellite map tile input $\mathcal{I}_{sat}$ with resized shape of $H \times W \times 3$, the satellite map branch adopts U-Net\cite{ronnebergerUNetConvolutionalNetworks2015} architecture with a ResNet18\cite{he2016deep} backbone to extract features $\mathcal{F}_{sat} \in \mathbb{R}^{H \times W\times C}$. The architecture is designed to maintain the structural integrity of the image and avoid potential distortions. Furthermore, we design a hierarchical fusion module to integrate $\mathcal{F}_{sat}$ and $\mathcal{F}_{bev}$. In the feature-level fusion stage, we decode a BEV-satellite attention mask derived from distance mask and segmentation mask, which is generated from $\mathcal{F}_{sat}$. Subsequently, a masked cross-attention is used to refine $\mathcal{F}_{bev}$ and produce refined features $\mathcal{F}_{ref}$. The BEV-level fusion introduces a BEV alignment module to ensure the alignment of $\mathcal{F}_{ref}$ and $\mathcal{F}_{sat}$ before concatenation and obtain the final fused features $\mathcal{F}_{fus}$. Lastly, a task-specific head is utilized for different tasks, including map semantic segmentation and instance detection.

\subsection{Feature-Level Fusion}

The feature-level fusion is designed to leverage information from satellite maps to enhance the features extracted from onboard sensors pixel by pixel. Given that the satellite maps cannot be updated in real-time, it is essential to follow the rule of prioritizing BEV features during the feature-level fusion. Therefore, we design position embeddings to modify the fusion weights of different positions and use masks to avoid interference from irrelevant information in satellite maps.

% linear embedding后的结果
Concretely, the inputs of the fusion module contain $\mathcal{F}_{bev}$ from onboard sensors and  $\mathcal{F}_{sat}$ from satellite maps. We perform patch embedding on the features using patches with the size of (5, 5) and enhance them with learnable positional embeddings. Linear projection is used to learn BEV features as queries $Q \in \mathbb{R}^{N \times C_h}$ and satellite map features as keys and values $K, V \in \mathbb{R}^{M \times C_h}$, where $N=800$, $M=800$, and $C_h=256$. We decode an attention mask $\mathcal{M}$, which is generated based on the segmentation of the satellite maps and distance information. Subsequently, the BEV features from onboard sensors are refined with the satellite map features by a masked cross-attention mechanism. Finally, the output is expanded and transformed to the BEV feature space to obtain the refined BEV features $\mathcal{F}_{ref}$.

We detail the position embeddings, mask generator, and masked cross-attention as follows.

% 有一点丑\mathbb{R}^{H_{grid} \times W_{grid} \times C}
\noindent \textbf{Position Embedding.} Vehicle onboard sensors are influenced by detection distances, whereas satellite maps are less susceptible to such influences. To achieve a balanced fusion process, we introduce position embeddings that emphasize the role of onboard sensors at close distances and satellite maps at longer distances. Specifically, we introduce two learnable parameters $PE_{bev} \in \mathbb{R}^{H_{grid} \times W_{grid} \times C}$ for BEV features from onboard sensor and $PE_{sat} \in \mathbb{R}^{H_{grid} \times W_{grid} \times C}$ for features from satellite maps, respectively. Here, $H_{grid}$ and $W_{grid}$ represent the height and width of patched BEV features. 

\noindent \textbf{Mask Generator.} Satellite maps may introduce irrelevant or erroneous information when they are obstructed or not updated in real-time, which can potentially interfere with BEV features from onboard sensors. Additionally, interactions from long distances have proved ineffective\cite{pang2023transcar}. To avoid unnecessary interactions between feature pairs, we introduce a BEV-Satellite attention mask $\mathcal{M} \in \mathbb{R}^{N \times M}$ that comprises distance mask and segmentation mask.

The distance mask is defined as,

\begin{equation}
\mathcal{M}_{dis}(x,y)=\begin{cases}
  -inf & \text{ if } {\rm Eul}(x,y)>D \\
  0 & \text{ else }
\end{cases} 
,
\end{equation}

\noindent where ${\rm Eul}(*)$ represents the Euclidean distance function, and $D=5$ represents the threshold of distance. 

Regarding the segmentation mask, we utilize a full convolution layer to generate the semantic segmentation information $\mathcal{S} \in \mathbb{R}^{H \times W}$ from $\mathcal{F}_{sat}$. Following this, we use patch embedding and linear embedding similar to BEV features to transform it into $\mathcal{E}_{seg} \in \mathbb{R}^{M \times 1}$ and expanded its dimensions to obtain the mask $\mathcal{M}_{seg} \in \mathbb{R}^{M \times N}$. The resultant mask $\mathcal{M}$ is defined as,

\begin{equation}
\mathcal{M} = \mathcal{M}_{seg}^{T} + \mathcal{M}_{dis},
\end{equation}

\noindent \textbf{Masked Cross-Attention.} Three masked cross-attention modules are cascaded to learn the associations between the BEV features from onboard sensors and features from satellite maps. We use the BEV features after linear embedding as queries $Q$ and the satellite map features as keys $K$ and values $V$, where $Q \in \mathbb{R}^{N \times C_h}$ and $K, V \in \mathbb{R}^{M \times C_h}$. Following \cite{chengMaskedattentionMaskTransformer2022}, the masked attention modulates the attention matrix via,

\begin{equation}
X = {\rm softmax}(\mathcal{M} + QK^{T})V + Q ,
\end{equation}

Finally, a feed-forward network (FFN) is used to calculate the refined features $Q_{out}$, which are of the same shape as the initial queries $Q$,

\begin{equation}
Q_{out} = {\rm FFN}(X) + X ,
\end{equation}

\subsection{BEV-Level Fusion}

Although the coordinate of satellite map has been coarsely calibrated during dataset generation, there may still be some discrepancies between the satellite maps and the actually generated BEV features considering the localization errors. Therefore, it is not appropriate to concaten the refined BEV feature $ \mathcal{F}_{ref}$ from vehicle sensors and features $\mathcal{F}_{sat}$ from satellite maps directly in BEV-level. To better fuse the two types of features, we perform an alignment operation before concatenation like the flow model\cite{huangAlignSegFeatureAlignedSegmentation2022, li2020semantic}. Specifically, $\mathcal{F}_{ref}$ and $\mathcal{F}_{sat}$ are concatenated together and passed through several convolutional layers to predict the coordinate offsets $\Delta \in \mathbb{R} ^ {H \times W \times 2}$ in each position. The warp operation is utilized to obtain aligned satellite map features $\tilde{\mathcal{F}}_{sat}$ by the bilinear interpolation kernel as,

\begin{equation}
    \begin{split}
        \tilde{\mathcal{F}}_{sat}(h, w) = \sum_{h'}^{H} \sum_{w'}^{W} \mathcal{F}_{h'w'} &\cdot \max (0,1-\left | h + \Delta _{hw1} - h' \right | ) \\
        & \cdot \max (0,1-\left | w + \Delta _{hw2} - w' \right | ),
    \end{split}
\end{equation}

% 加图

\noindent where $\Delta_{hw1}$, $\Delta_{hw2}$ represent the learned 2D transformation offsets for position $(h, w)$. The addition operation involves adding $\mathcal{F}_{sat}$ to $\tilde{\mathcal{F}}_{sat}$. We then concatenate $\tilde{\mathcal{F}}_{sat}$ and $\mathcal{F}_{ref}$ to obtain the final fused features $\mathcal{F}_{fus}$. Finally, we utilize the task-specific head to construct the HD maps.

\section{Experiments}

\subsection{Dataset \& Metrics}

We evaluate our module on nuScenes and satellite map complementary dataset. We focus on two sub-tasks in HD map construction: map semantic segmentation and instance detection. We utilize HDMapNet as a baseline for the map semantic segmentation task and evaluate the performance using Mean Intersection over Union (mIoU). For the map instance detection task, we use MapTR and VectorMapNet as the baselines and evaluate the performance using Mean Average Precision (mAP). 

\subsection{Baseline Models}

To validate the broad applicability of our methods, we incorporate our fusion module into three recently proposed camera-based HD map construction methods, which serve as our baseline methods as follows:

\noindent \textbf{HDMapNet}\cite{liHdmapnetOnlineHd2022} introduces the map learning problem. It utilizes the MLP-based projection method to extract BEV features and involves a post-processing step to generate vectorized map elements. Apart from comparative experiments, we perform long-range construction experiments and ablation studies using HDMapNet as a baseline.

\noindent \textbf{VectorMapNet}\cite{liuVectormapnetEndtoendVectorized2022} is the pioneering work in end-to-end map instance detection. It adopts two Transformers to predict key points and generate map elements, respectively.

\noindent \textbf{MapTR}\cite{liaoMapTRStructuredModeling2022} proposes a novel modeling approach for map elements and achieves the current state-of-the-art performance.

\subsection{Comparison with Baselines}

\captionsetup[table]{singlelinecheck=false}

\noindent \textbf{Effectiveness of satellite maps}.  We integrate the satellite map fusion module into three distinct baselines, each possessing varying network architectures, perspective-to-BEV projection methods, and construction tasks. Table \ref{table-1} and Table \ref{table-2} show the comparisons. In the map segmentation task, the satellite map fusion module achieves 20.8 higher IoU compared to HDMapNet. In the map instance detection task, it also achieves 7.9 higher mAP compared to VectorMapNet. Since MapTR implicitly represents BEV features, we perform patch embedding on the satellite map features and then utilize deformable attention for feature-level fusion. Surprisingly, this approach still achieves 2.3 higher mAP. These findings indicate that our proposed satellite map fusion module is a general approach that can potentially be applied to other HD map construction frameworks.

% It's worth mentioning that our performance improvement was achieved within the same epochs as the baseline model, and the fusion module does not slow down the convergence. 

\noindent \textbf{Influence of detection range}. The detection range of HD maps greatly influences downstream planning and decision-making modules. However, due to the inherent limitations of onboard sensors, the performance of models declines significantly as the detection range increases. Compared to onboard sensors, satellite maps can provide information from long-range detection. We set various BEV ranges for comparison, including $ 60m \times 30m $, $60m \times 60m$, and $120m \times 60m$. The enhancement is shown in Table \ref{table-3}. After integrating the satellite map fusion module, we observe that as the distance increases, the rate of performance degradation of the model significantly decreases. 

\begin{table}[htb]
\centering
% \vspace{-1em}
\caption{IoU scores (\%) of map semantic segmentation on the nuScenes validation set. SFM stands for the satellite map fusion module. By adding satellite map information, the fusion module can improve the performance of HDMapNet(HDMapNet remains the same as in the original work).}
\label{table-1}
\begin{tabular}{cc|cccc}
\toprule
\multirow{2}{*}{Baseline} & \multirow{2}{*}{+ SFM} & \multicolumn{4}{c}{IoU score (\%)} \\
 & & Divider & Crossing & Boundary & All \\
\midrule
\multirow{2}{*}{HDMapNet} & - & 40.6 & 18.7 & 39.5 & 32.9 \\
 & $\checkmark$ & 54.9 & 53.4 & 52.9 & 53.7 \\
\cmidrule(lr){1-2} \cmidrule(l){3-6}
\multicolumn{2}{c|}{$\Delta mIoU$} & \textbf{+14.3} & \textbf{+34.7} & \textbf{+13.4} & \textbf{+20.8} \\
\bottomrule
\end{tabular}
\vspace{-1em}
\end{table}

\begin{table}[htb]
\centering
\caption{mAP of map instance detection on the nuScenes validation set. By adding satellite map information, the fusion module boosts the performance of both VectorMapNet and MapTR(VectorMapNet and MapTR remain the same as in the original work).}
\label{table-2}
\begin{tabular}{cc|cccc}
\toprule
\multirow{2}{*}{Baseline} & \multirow{2}{*}{+ SFM} & \multicolumn{4}{c}{mAP (\%)} \\
 & & Divider & Crossing & Boundary & All \\
\midrule
\multirow{2}{*}{VectorMapNet} & - & 47.3 & 36.1 & 39.3 & 40.9 \\
 & $\checkmark$ & 51.9 & 50.2 & 44.2 & 48.8 \\
\cmidrule(lr){1-2} \cmidrule(lr){3-6}
\multicolumn{2}{c|}{$\Delta mAP$} & \textbf{+4.6} & \textbf{+14.1} & \textbf{+4.9} & \textbf{+7.9} \\
\midrule
\multirow{2}{*}{MapTR} & - & 51.5 & 46.3 & 53.1 & 50.3 \\
 & $\checkmark$ & 55.3 & 47.2 & 55.3 & 52.6 \\
\cmidrule(lr){1-2} \cmidrule(lr){3-6}
\multicolumn{2}{c|}{$\Delta mAP$} & \textbf{+3.8} & \textbf{+0.9} & \textbf{+2.2} & \textbf{+2.3} \\
\bottomrule
\end{tabular}
\vspace{-1em}
\end{table}

\begin{table}[ht!]
\centering
\caption{Comparison of model performance using HDMapNet as the baseline at different BEV ranges. }
\label{table-3}
\begin{tabular}{cc|cccc}
\toprule
\multirow{2}{*}{BEV Range} & \multirow{2}{*}{+ SFM} & \multicolumn{4}{c}{IoU score (\%)} \\
 & & Divider & Crossing & Boundary & All \\
\midrule
\multirow{2}{*}{$60m \times 30m$} & - & 40.6 & 18.7 & 39.5 & 32.9 \\
 & $\checkmark$ & 54.9 & 53.4 & 52.9 & 53.7 \\
\cmidrule(lr){1-2} \cmidrule(lr){3-6}
\multicolumn{2}{c|}{$\Delta mIoU$} & \textbf{+14.3} & \textbf{+34.7} & \textbf{+13.4} & \textbf{+20.8}  \\
\midrule
\multirow{2}{*}{$60m \times 60m$} & - & 33.6 & 15.8 & 32.2 & 27.2 \\
 & $\checkmark$ & 51.6 & 52.1 & 49.1 & 50.9 \\
\cmidrule(lr){1-2} \cmidrule(lr){3-6}
\multicolumn{2}{c|}{$\Delta mIoU$} & \textbf{+18.0} & \textbf{+36.3} & \textbf{+16.9} & \textbf{+23.7} \\
\midrule
\multirow{2}{*}{$120m \times 60m$} & - & 26.9 & 12.9 & 25.7 & 21.8 \\
 & $\checkmark$ & 51.0 & 53.0 & 45.2 & 49.7 \\
\cmidrule(lr){1-2} \cmidrule(lr){3-6}
\multicolumn{2}{c|}{$\Delta mIoU$} & \textbf{+24.1} & \textbf{+40.1} & \textbf{+19.5} & \textbf{+27.9} \\
\bottomrule
\end{tabular}
\vspace{-1em}
\end{table}

\begin{figure*}[t]

    \centering
    \includegraphics[width=1.9\columnwidth]{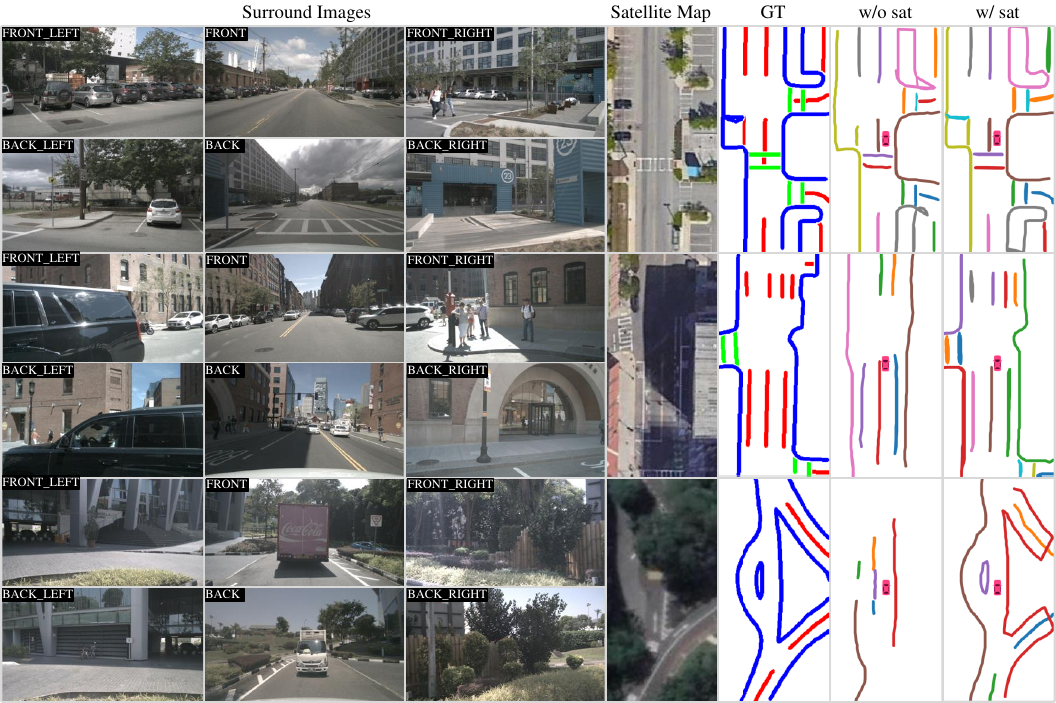}
    \caption{Qualitative results of our method, where sat stands for satellite maps and GT stands for ground truth. After incorporating satellite maps, the model's performance is significantly improved in both complex scenarios and situations of occlusion by other vehicles. Moreover, the model exhibits stable enhancement in areas not covered by satellite maps.}
    \label{visual}
    \vspace{-1em}
    
\end{figure*}

\subsection{Ablation Studies}

\noindent \textbf{Feature-level fusion module}. In Table \ref{table-4}, we conduct ablation studies on different feature fusion methods, which include concatenation directly, standard attention\cite{vaswaniAttentionAllYou2017}, deformable attention\cite{zhuDeformableDetrDeformable2020}, shift window attention\cite{liuSwinTransformerHierarchical2021}, and masked attention(ours). We observed that direct concatenation leads to a performance improvement, demonstrating that satellite maps can effectively complement onboard sensors in HD map construction. On the other hand, attention-based methods show further enhancement in fusion performance. Notably, our proposed distance and segmentation mask strategies can achieve 1.6 higher mIoU compared to standard attention. This improvement can be attributed to the mask's ability to reduce unnecessary interactions, particularly in minor scenarios where satellite maps contain errors or irrelevant information.

\begin{table}[t]
\centering
\caption{Ablation on the feature-level fusion module using HDMapNet as baseline. Concat stands for concatenation. SA stands for standard attention\cite{vaswaniAttentionAllYou2017}. DA stands for deformable attention\cite{zhuDeformableDetrDeformable2020}. SWA stands for shift window attention\cite{liuSwinTransformerHierarchical2021}. MA(d) stands for distance mask attention. MA(d+s) stands for distance and segmentation mask attention.}
\label{table-4}
\begin{tabular}{c|cccc}
\toprule
\multirow{2}{*}{Method} & \multicolumn{4}{c}{IoU score (\%)} \\
 & Divider & Crossing & Boundary & All \\
\midrule
baseline & 40.6 & 18.7 & 39.5 & 32.9 \\
concat & 53.6(+13.0) & 45.2(+26.5) & 46.0(+6.5) & 48.3(+15.4) \\
SA & 51.5(+10.9) & \textbf{51.5(+32.8)} & 50.2(+10.7) & 51.1(+18.2) \\
DA & 53.4(+12.8) & 50.3(+31.6) & 51.5(+12.0) & 51.7(+18.8) \\
SWA & 54.0(+13.4) & 49.6(+30.9) & 52.0(+12.5) & 51.9(+19.0) \\
\textbf{MA(d)}& 53.4(+12.8) & 50.2(+31.5) & 51.0(+11.5) & 51.5(+18.6) \\
\textbf{MA(d+s)} & \textbf{54.5(+13.9)} & \textbf{51.5(+32.8)} & \textbf{52.2(+12.7)} & \textbf{52.7(+19.8)} \\
\bottomrule
\end{tabular}
\vspace{-2em}
\end{table}

\noindent \textbf{BEV-level fusion module}. Ablations on the BEV-level fusion are presented in Table \ref{table-5}. The experimental results indicate that solely performing coarse alignment during satellite map slicing leads to a decrease in the model's performance due to coordinate deviations. By introducing an alignment module before concatenation, the impact of coordinate deviations can be effectively alleviated, resulting in an improvement of 1.0 mIoU.

\subsection{Qualitative Visualization}
We visualize the HD map construction in Fig. \ref{visual}. The first scenario illustrates that the satellite map can provide additional information and enhance the model capability in complex scenes, such as intersections. In the second scenario, the vehicle's left-side field of view is obstructed by another vehicle, resulting in the omission of an intersection. The satellite map capitalizes on its long-range data provision capability to aid in intersection detection. In the final scenario, where the satellite map is obstructed, our proposed method still consistently improves the model's performance, and this can be attributed to the mask's ability to avoid interference from irrelevant information. 

\begin{table}[t]
\centering
\caption{Ablation on the BEV-level fusion using HDMapNet as the baseline.}
\label{table-5}
\begin{tabular}{c|cccc}
\toprule
\multirow{2}{*}{Method} & \multicolumn{4}{c}{IoU score (\%)} \\
 & Divider & Crossing & Boundary & All \\
\midrule
baseline & 40.6 & 18.7 & 39.5 & 32.9 \\
w/o align & 54.5(+13.9) & 51.5(+32.8) & 52.2(+12.7) & 52.7(+19.8) \\
alignment & \textbf{54.9(+14.3)} & \textbf{53.4(+34.7)} & \textbf{52.9(+13.4)} & \textbf{53.7(+20.8)} \\
\bottomrule
\end{tabular}
\vspace{-2em}
\end{table}

\section{Conclusion}

In this paper, we explore the use of cloud-based satellite maps to complement onboard sensors for boosting HD map construction. We generate the corresponding satellite map tiles for each sample in nuScenes and release them as a complementary dataset. To integrate the satellite maps into existing methods, we propose a hierarchical fusion module that enhances features obtained from onboard sensors at the feature level and aligns features at the BEV level. Comprehensive experiments demonstrate that satellite map information and our proposed method can enhance the performance of existing models, particularly in long-range map construction. We believe that combining cloud-based data such as satellite maps with onboard sensor data will offer a novel perspective to future HD map construction tasks.

%%%%%%%%%%%%%%%%%%%%%%%%%%%%%%%%%%%%%%%%%%%%%%%%%%%%%%%%%%%%%%%%%%%%%%%%%%%%%%%%

%%%%%%%%%%%%%%%%%%%%%%%%%%%%%%%%%%%%%%%%%%%%%%%%%%%%%%%%%%%%%%%%%%%%%%%%%%%%%%%%



% \section*{ACKNOWLEDGMENT}

% The preferred spelling of the word ÒacknowledgmentÓ in America is without an ÒeÓ after the ÒgÓ. Avoid the stilted expression, ÒOne of us (R. B. G.) thanks . . .Ó  Instead, try ÒR. B. G. thanksÓ. Put sponsor acknowledgments in the unnumbered footnote on the first page.

%%%%%%%%%%%%%%%%%%%%%%%%%%%%%%%%%%%%%%%%%%%%%%%%%%%%%%%%%%%%%%%%%%%%%%%%%%%%%%%%

% References are important to the reader; therefore, each citation must be complete and correct. If at all possible, references should be commonly available publications.

\bibliographystyle{IEEEtran}

% \bibliography{IEEEexample}
\bibliography{paper1}

\end{document}